\documentclass[runningheads]{llncs}

\usepackage{eccv}

\usepackage{eccvabbrv}

\usepackage{graphicx}
\usepackage{booktabs}

\usepackage[accsupp]{axessibility}  
\usepackage{multirow}
\usepackage{bm}

\usepackage{hyperref}
\usepackage[capitalize]{cleveref}
\usepackage{orcidlink}

\begin{document}

\newcommand{\red}[1]{{\color{red}#1}}
\newcommand{\todo}[1]{{\color{red}#1}}
\newcommand{\TODO}[1]{\textbf{\color{red}[TODO: #1]}}

\newcommand{\methodName}{\emph{HairOrbit}\xspace}







\title{\methodName: Multi-view Aware 3D Hair Modeling from Single Portraits} 

\titlerunning{HairOrbit}

\author{Leyang Jin\inst{1} \and
Yujian Zheng\inst{1}\thanks{Corresponding author.}  \and
Bingkui Tong\inst{1} \and
Yuda Qiu\inst{2} \and
Zhenyu Xie\inst{1} \and
Hao Li\inst{1,3}
}

\authorrunning{L.~Jin et al.}

\institute{Mohamed bin Zayed University of Artificial Intelligence \and
SSE, The Chinese University of Hong Kong, Shenzhen
\\
\and
Pinscreen\\
}

\maketitle

\begin{abstract}
Reconstructing strand-level 3D hair from a single-view image is highly challenging, especially when preserving consistent and realistic attributes in unseen regions. Existing methods rely on limited frontal-view cues and small-scale/style-restricted synthetic data, often failing to produce satisfactory results in invisible regions.
In this work, we propose a novel framework that leverages the strong 3D priors of video generation models to transform single-view hair reconstruction into a calibrated multi-view reconstruction task.
To balance reconstruction quality and efficiency for the reformulated multi-view task, we further introduce a neural orientation extractor trained on sparse real-image annotations for better full-view orientation estimation.
In addition, we design a two-stage strand-growing algorithm based on a hybrid implicit field to synthesize the 3D strand curves with fine-grained details at a relatively fast speed.
Extensive experiments demonstrate that our method achieves state-of-the-art performance on single-view 3D hair strand reconstruction on a diverse range of hair portraits in both visible and invisible regions.
\end{abstract}

\section{Introduction}
\label{sec:intro}
Reconstructing strand-level 3D hair geometry from a single image remains one of the most challenging tasks in digital human modeling. 
Hair exhibits highly complex structures with self-occlusions, intricate topology, and diverse local details such as curliness, partition, and varying lengths. 
In single-view reconstruction, which is an inherently ill-posed problem, the visible regions provide only partial cues, while the occluded parts must be inferred from learned priors that ensure both geometric plausibility and stylistic consistency. 
Achieving such coherence between visible and invisible regions is essential for realistic digital avatars and AR/VR applications, yet it remains an open challenge.


\begin{figure}
	\centering
	\includegraphics[width=\linewidth]{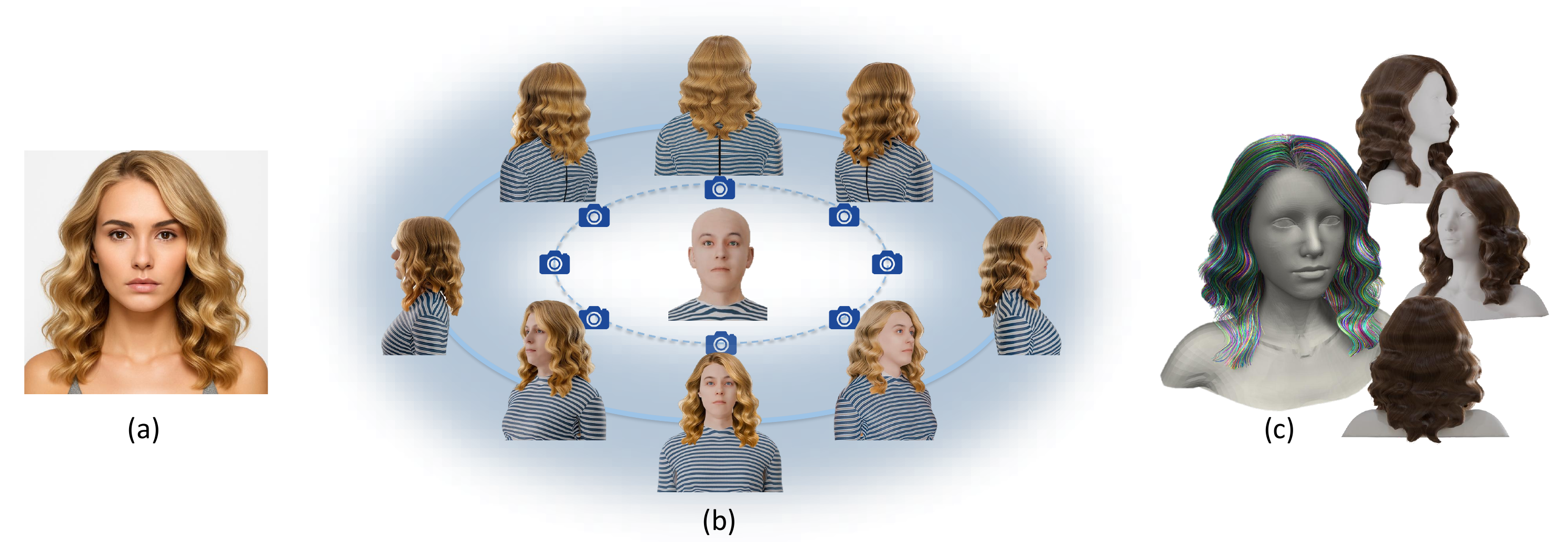}
	\caption{We propose a novel framework for strand-level single-view 3D hair reconstruction. Given a frontal-view portrait (a), we first synthesize corresponding calibrated multi-view images (b) on a camera orbit, then reconstruct multi-view aware 3D hair strands (c). Note that the left view in (c) is rendered with about 10k strands to better visualize the geometry, while the other 3 views are rendered with 100k.}
    \label{fig:teaser}
    \vspace{-2em}
\end{figure}

Existing single-view methods often rely on limited appearance cues and small-scale or style-restricted 3D hair datasets~\cite{sklyarova2025im2haircut, zheng2023hairstep, rosu2025difflocks,wu2022neuralhdhair} to hallucinate occluded geometry.
However, due to the scarcity and limited realism of current synthetic 3D hair datasets~\cite{rosu2025difflocks, zhou2018hairnet}, these priors are insufficient to capture the diversity and structural coherence of real-world hairstyles, leading to unnatural, over-smoothed, or image-unaligned reconstructions.
Notably, a non-negligible limitation of these methods is that the reconstructed hair often appears plausible from the frontal view but fails to maintain style consistency or suffers from severe artifacts under unseen viewpoints.
In contrast, recent state-of-the-art multi-view approaches~\cite{wu2024monohair, zhou2024groomcap, zakharov2024human,sklyarova2023neural,kuang2022deepmvshair} demonstrate impressive performance in capturing high-fidelity, strand-level details, as they can leverage fine and geometrically consistent information from captured multi-view images.
This raises a natural question: can multi-view information be synthesized to assist single-view reconstruction?

Inspired by recent advancements in video diffusion models~\cite{wan2025wan}, we observe that such models intrinsically encode strong 3D priors through their temporal consistency.
Building on this insight, we combine the priors learned from large-scale video generation models with a limited but expertly curated set of high-quality synthetic 3D hair data~\cite{hu2015single,rosu2025difflocks}, to validate our hypothesis that single-view hair reconstruction can be reformulated as a multi-view task.
Specifically, we fine-tune a video diffusion model using a lightweight LoRA~\cite{hu2022lora} on a carefully selected subset of 3D hair models and their corresponding multi-view renderings, enabling the generation of consistent video sequences in a 360° orbital rotation from a single input image.
By leveraging these synthesized multi-view sequences as an intermediate bridge, we effectively transform the single-view reconstruction problem into a calibrated multi-view setting, allowing us to exploit the advantages of multi-view methods without requiring captured real multi-view data (\cref{fig:teaser}).

However, when revisiting multi-view hair strand reconstruction, several issues remain.
The most critical one lies in the dependency of current methods on orientation maps extracted using Gabor filters~\cite{paris2004capture}.
Such maps are typically noisy and inaccurate, which severely affects the geometric consistency across views.
Existing approaches either rely on time-consuming denoising~\cite{wu2024monohair} or optimization processes~\cite{zakharov2024human} to achieve satisfactory results, or suffer from degraded quality~\cite{kuang2022deepmvshair}.
To address this, we train a neural orientation extractor using manually annotated hair-direction data from~\cite{zheng2023hairstep}, which produces accurate, clean, and smooth orientation maps.
Unlike~\cite{zheng2023hairstep}, which aims to predict a directed global orientation that requires semantic understanding and therefore struggles to generalize to back or side views when only trained on frontal views, we observe that orientation extraction is essentially a local, pixel-aligned task.
Consequently, by training on annotated frontal-view orientations, our model generalizes well to full-view cases.
The effectiveness of our method is validated on a newly annotated dataset of 395 images with hair growth strokes similar to~\cite{zheng2023hairstep}, but covering full-view cases captured from multi-view videos~\cite{sklyarova2023neural, wu2024monohair}.

Following~\cite{wu2024monohair, kuang2022deepmvshair, wu2022neuralhdhair, zheng2023hairstep}, we predict an implicit 3D hair field to guide strand growth based on the aggregated features from multi-view orientation maps and estimated depth maps.
This leads to the second question concerning the implicit field representation and the hair-growing mechanism.
Existing methods typically construct two separate fields: an orientation field and an occupancy field.
We find this occupancy field actually inefficient for hair growing. Instead, we introduce a hybrid field that jointly models both behaviors: within the hair volume, it learns the orientation field, and outside, it predicts zero-valued growth vectors.
With this formulation, hair strands naturally stop when reaching the boundary, and repeated growth steps always converge to the same endpoints.
This property enables highly efficient parallel strand growth, leading to a substantial speedup.
We combine scalp-root and segment-based growth to achieve fine and efficient strand generation~\cite{chai2013dynamic}, avoiding the incompleteness of HairStep and the high cost of MonoHair.
In addition, we filter out redundant or collapsed short strands brought by pre-sampled roots and recover collapsed buzz-cut using a re-projection method.

The main contributions of our work are as follows:
\begin{itemize}
    \item We reformulate single-view hair reconstruction as a calibrated multi-view reconstruction problem by leveraging the intrinsic cross-view priors embedded in video diffusion models.
    \item We propose a multi-view neural orientation extractor, a hybrid implicit field, and a combined strand-growing algorithm to effectively balance reconstruction fidelity and efficiency for more practical applications.
    \item We achieve state-of-the-art performance in single-view strand-level 3D hair reconstruction, simultaneously maintaining visual alignment with the input image and ensuring plausible multi-view geometry. 
\end{itemize}

\section{Related Work}
\label{sec:related}

\paragraph{Strand-level 3D hair modeling}
How to acquire human hair in 3D has been an active research topic for decades.
Multi-view stereo methods have long been explored to reconstruct 3D hair geometry from multiple captured images~\cite{paris2008hair, luo2012multi, luo2013wide, hu2014robust, nam2019strand, zhou2024groomcap}.
To reduce the reliance on specialized equipment and constrained capture setups, several approaches attempt to model hair from less restricted inputs, such as sparse-view images~\cite{zhang2017data}, selfie videos~\cite{liang2018video}, and RGB-D streams~\cite{zhang2018modeling}.
More recently, high-quality reconstructions from monocular videos have been demonstrated~\cite{wu2024monohair, sklyarova2023neural}.
However, these methods require long processing times (e.g., 3–4 days in~\cite{sklyarova2023neural} and 4–6 hours in~\cite{wu2024monohair}), limiting their accessibility to ordinary users.

In contrast, several studies focus on reconstructing 3D hair from a single-view image, which are more efficient.
Early methods~\cite{chai2012single,chai2013dynamic} lift partial strands from 2D to manipulate portraits.
The pioneering retrieval-based methods~\cite{hu2015single,chai2016autohair} typically retrieve a coarse hair model from a database and then refine it through geometric optimization. 
The effectiveness of these approaches relies on the quality of priors, and the performance is less satisfactory for challenging input.
With the rise of deep learning, data-driven methods~\cite{zhou2018hairnet, saito20183d, shen2020deepsketchhair, wu2022neuralhdhair, zheng2023hairstep} have emerged, leveraging deep neural networks for reconstruction.
However, these methods often produce coarse or over-smoothed geometry due to the limited prior from the scale, diversity, and quality of 3D synthetic data. 
Recent approaches~\cite{rosu2025difflocks, sklyarova2024text} further generate hair texture maps in a latent space, though such representations are usually misaligned with the input view.
Differentiable-rendering-based methods~\cite{sklyarova2025im2haircut, tang2025single} alleviate some limitations on the alignment but still remain slow and struggle to recover unseen back-view regions.

\paragraph{Orientation maps for hair modeling}
The intricate and tangled structure of human hair makes directional cues an intuitive foundation for 3D reconstruction.
Instead of modeling hair geometry directly, many approaches first infer intermediate orientation representations that describe strand flow in either 2D or 3D space.
Classical image-based methods estimate 2D orientation maps by convolving the input portrait with a bank of Gabor filters and taking the direction that yields the strongest response~\cite{paris2004capture, paris2008hair}.
Such orientation maps can then be lifted into 3D through multi-view calibration~\cite{luo2012multi, luo2013wide, hu2014robust} or supplied to neural networks as auxiliary guidance for predicting volumetric hair structures~\cite{zhou2018hairnet, wu2022neuralhdhair, yang2019dynamic, zhang2019hair}.
However, 2D orientations obtained from local filtering are highly sensitive to noise, which can be mitigated via an additional postprocessing process~\cite{luo2012multi, luo2013wide, wu2024monohair}.
~\cite{zheng2023hairstep} introduces a neural orientation extractor that predicts directional strand maps directly from raw images to resolve growing-direction ambiguity.
However, it is limited to frontal-view inputs.

\paragraph{3D hair dataset}
Lack of large-scale and high-quality 3D data remains a major bottleneck in hair modeling research.
Currently, USC-HairSalon~\cite{hu2015single} is the most widely used open-source dataset, containing 343 hair models across 12 unbalanced categories.
\cite{zhou2018hairnet} augments the data from~\cite{hu2015single} by randomly combining parts from hair strand models with similar styles, but the resulting data are still limited in both realism and diversity.
Besides, 3DHW~\cite{chai2016autohair} collects 50K portrait images and reconstructs hair geometry through traditional single-view optimization. 
Due to the inherent limitations of this algorithm, the resulting 3D models often lack fidelity and structural accuracy.
~\cite{zheng2024towards} proposes a large-scale dataset of calibrated multi-view images with 2,396 hairstyles, but without corresponding 3D strands, which limits its applicability to strand-level reconstruction tasks.
Recently, ~\cite{rosu2025difflocks} builds a dataset of 40K 3D hairstyles with rendered images, yet the variation remains limited, with many repetitive styles and unnatural poses.
Since creating high-quality 3D hair models at scale is costly, we seek to move away from heavy reliance on such data and leverage multi-view priors to improve the generalization of single-view reconstruction.


\section{Method}

\subsection{Overview}
\label{subsec:overview}
\begin{figure}
	\centering
	\includegraphics[width=1.0\linewidth]{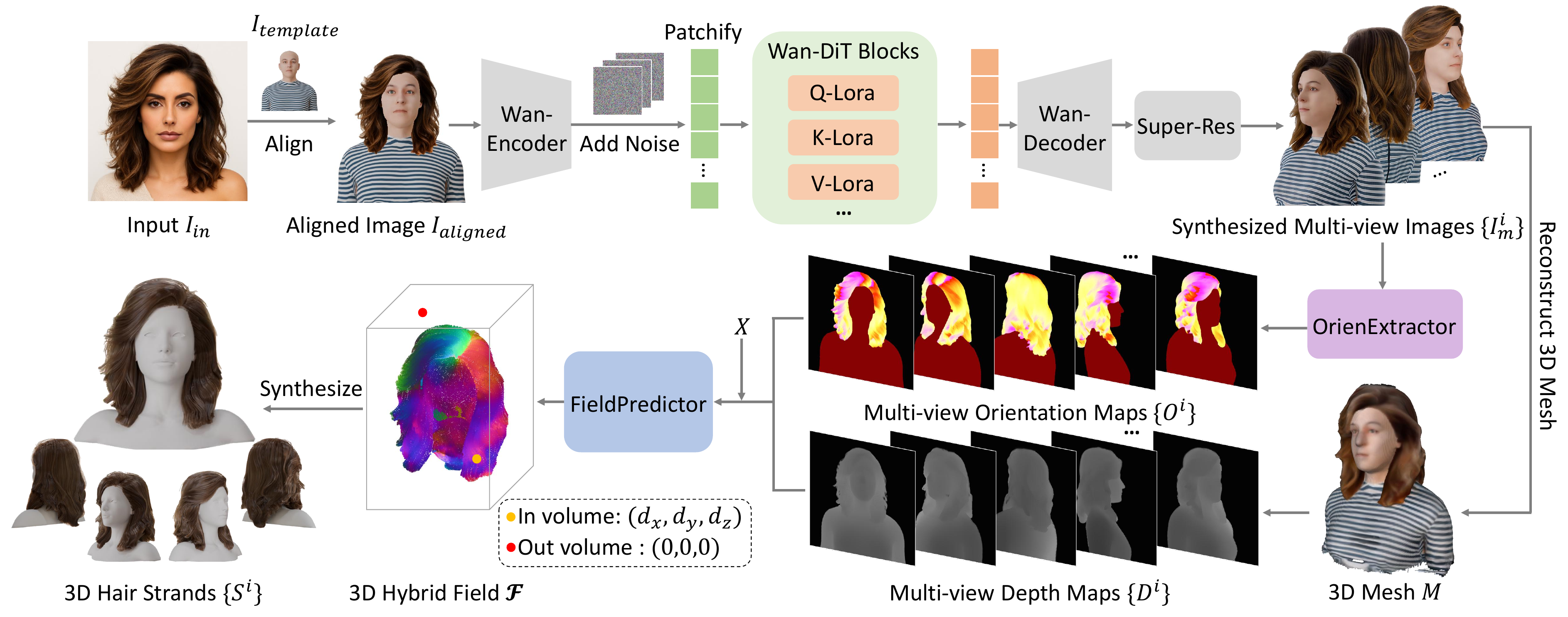}
	\caption{Overview of \methodName. Given a single portrait, \methodName converts single-view 3D hair reconstruction into a multi-view task.}
    \label{fig:pipeline}
    \vspace{-3em}
\end{figure}

Aiming to improve the style consistency and overall quality of single-view hair reconstruction, we propose a novel pipeline, \methodName, as illustrated in~\cref{fig:pipeline}.
Given a frontal-view portrait $I_{in}$, we first align the hair region with a template body image $I_{template}$ to obtain $I_{aligned}$ by fitting landmarks as~\cite{zheng2024towards}.
From $I_{aligned}$, we generate calibrated multi-view images $\{I_m^i\}$ using a video diffusion model (see~\cref{subsec:mvgen}).
Based on $\{I_m^i\}$, we reconstruct a 3D mesh $M$ and render the corresponding multi-view depth maps $\{D^i\}$.
In parallel, multi-view orientation maps $\{O^i\}$ are predicted from $\{I_m^i\}$.
Subsequently, we predict a hybrid implicit field $\bm{\mathcal{F}}$ from the aggregated features of $\{O^i\}$ and $\{D^i\}$, to synthesize 3D hair strands $\{S^i\}$ (see~\cref{subsec:3drecon}).

\subsection{Multi-view Generation}
\label{subsec:mvgen}

Given the aligned image $I_{aligned}$, the goal of multi-view generation is to synthesize calibrated multi-view images $\{I_m^i\}$ along a predefined orbital path.
We leverage the intrinsic cross-view priors embedded in video diffusion models, and adapt the model to our setting by training a LoRA~\cite{hu2022lora} on a small set of synthetic multi-view renderings of 3D hairstyles.

\paragraph{Video diffusion model.}
To ensure temporal consistency in the generated results, we adopt WAN~\cite{wan2025wan} as our backbone.
A core element of this framework is a VAE that encodes video frames into a temporally coherent latent representation, ensuring causal consistency while maintaining computational efficiency.
In this latent space, a Diffusion Transformer (DiT) is employed to generate latent representations of multi-view 360° rotated video frames from noise, while conditioning on the latent derived from our aligned input image.
The generated latents are then decoded into multi-view frames.
To accommodate consumer-grade GPU hardware constraints, we employ a lower-resolution version of the model during inference, which leads to synthesized multi-view images that are not sufficiently sharp for accurate orientation extraction.
To address this, we enhance the generated multi-view images using the Flux.1-dev Upscaler~\footnote{\url{https://huggingface.co/jasperai/Flux.1-dev-Controlnet-Upscaler}}~\cite{flux2024,labs2025flux1kontextflowmatching}
, obtaining high-resolution results $\{I_m^i\}$ with improved hair texture and fine-grained details.

\paragraph{LoRA training for orbital hair rotation.}
To guide the video diffusion model toward our desired generation pattern using only a small amount of data, while minimizing the forgetting of its pre-learned real-world hair priors, we fine-tune the model with LoRA applied to the Q, K, V, O projections and the first and last linear layers (FFN.0 and FFN.2) of each transformer block.
We prepare multi-view renderings from carefully selected typical 3D hairstyles, covering a wide range of variations in length, curliness, and partition.
Following diffusion-based~\cite{rombach2022high} video generation models, we optimize the LoRA parameters using a standard noise prediction loss.
Given a clean latent $\mathbf{x}_0$ and a timestep $t \!\sim\! \mathcal{U}(1, T)$, a noisy latent $\mathbf{x}_t$ is obtained by the forward diffusion process:
\begin{equation}
\mathbf{x}_t = \sqrt{\bar{\alpha}_t}\,\mathbf{x}_0 
+ \sqrt{1 - \bar{\alpha}_t}\,\boldsymbol{\epsilon},
\quad \boldsymbol{\epsilon} \sim \mathcal{N}(0, \mathbf{I}).
\end{equation}
where $\bar{\alpha}_t$ denotes the cumulative noise schedule.
The model $\epsilon_\theta(\mathbf{x}_t, t, \mathbf{c})$ predicts the noise $\boldsymbol{\epsilon}$ conditioned on $\mathbf{c}$,
which in our case corresponds to the latent representation of the aligned input image.
The training loss is defined as:
\begin{equation}
\mathcal{L}_{\text{denoise}}
= \mathbb{E}_{\mathbf{x}_0, \boldsymbol{\epsilon}, t}
\big[
\|\epsilon_\theta(\mathbf{x}_t, t, \mathbf{c})
- \boldsymbol{\epsilon}\|_2^2
\big],
\end{equation}
where $\mathbf{x}_0$ is the clean latent encoded by the VAE,
$\mathbf{x}_t$ is the noisy latent at step $t$,
$\bar{\alpha}_t$ is the cumulative product of noise coefficients,
$\boldsymbol{\epsilon}$ is the Gaussian noise to be predicted,
$\epsilon_\theta$ is the DiT backbone with LoRA adapters,
and $\mathbf{c}$ is the conditioning latent from the aligned image $I_{aligned}$.
During fine-tuning, only the LoRA weights are updated, while all pretrained parameters remain frozen.

\subsection{Hair Strands Reconstruction}
\label{subsec:3drecon}
After the previous stage, we successfully transfer the single-view reconstruction to a calibrated multi-view reconstruction problem.
Following \cite{wu2024monohair, kuang2022deepmvshair}, we adopt the paradigm that predicting an implicit field from multi-view orientation maps and depth maps.
Then we generate 3D hair strands according to the field value by certain strand growing algorithm.

\paragraph{Orientation Extraction}
Due to the inherent difficulty of realistic hair rendering, existing hair modeling approaches~\cite{wu2022neuralhdhair, chai2013dynamic, zhang2019hair, zhou2018hairnet, kuang2022deepmvshair, wu2024monohair} typically use orientation maps instead of raw hair images to reduce the domain gap between synthetic and real data. 
With pixels indicating 2D growing direction in local regions, we hypothesize that providing clean and smooth orientation maps $\{O^i\}$ can significantly benefit 3D reconstruction.
However, orientation maps obtained through image filtering are often contaminated by noise and thus fail to serve as reliable intermediate representations, leading to inaccuracies in 3D reconstruction.
~\cite{wu2024monohair} attempts to address this problem via patch-based multi-view optimization for denoising, but this makes the overall process extremely time-consuming.
~\cite{zheng2023hairstep} eliminates the reliance on Gabor filters by training a neural 2D orientation estimator on the HiSa dataset, which provides 2D directional curve annotations, allowing direct prediction of strand orientations from raw images.
However, HiSa contains only frontal-view annotations, and the directional strand maps used in~\cite{zheng2023hairstep} rely heavily on global context information. Thus, it often fails on side and back views.

We observe that orientation extraction is inherently a pixel-aligned and locally perceptual task.
We utilize frontal-view annotations in HiSa to learn orientation estimation across multiple viewpoints.
Following their design, we colorize SVG strand curves to generate training orientation maps within the hair mask region and train a U-Net~\cite{ronneberger2015u} model using a combination of $\ell_1$ and perceptual losses. 
Although we use the same network as~\cite{zheng2023hairstep}, changing the learning target from a directional strand map to an undirectional orientation map, makes our method more applicable and accurate for full-view tasks, rather than being limited to frontal views like HairStep~\cite{zheng2023hairstep}.

To verify the performance of the orientation extractor, we also annotate a set of multi-view images with SVG curves and conduct an evaluation on orientation prediction.
Please check ~\cref{subsec: evaluation} for details.

\paragraph{Depth estimation}
Since we have consistent multi-view images $\{I_m^i\}$ from the video diffusion model, it is straightforward for us to reconstruct a mesh $M$ and render depth maps $D^i$ for each frame $i$.
In practice, considering the time efficiency, we use these multi-view images to fit Gaussians~\cite{huang20242d}.
~\cite{huang20242d} collapses the 3D volume into a set of 2D oriented planar Gaussian disks, providing view-consistent geometry while modeling surfaces intrinsically.
After training, a surface mesh can be obtained via marching cubes~\cite{lorensen1998marching}, which will be rendered to multi-view depth.

\paragraph{Hybrid field reconstruction}
Existing methods typically construct two separate fields: an orientation field and an occupancy field.
The occupancy field defines hair boundaries to constrain growth, while the orientation field learns directional cues only within the hair volume and depends on the occupancy field to terminate strands outside.
We observe that the occupancy field is inefficient, introducing costly per-point queries and extra ray-casting operations after marching cubes in the hair-growing process.
Instead, we introduce a hybrid field $\bm{\mathcal{F}}$ that unifies both behaviors by learning orientations inside the hair volume and predicting zero growth vectors outside.
With this design, strands naturally stop at the boundary, and repeated growth steps consistently converge to the same endpoints, enabling efficient parallel strand growth and yielding a significant speedup.

The implicit function of $\bm{\mathcal{F}}$ is formulated as:
\begin{equation}
F(X, (O^1, D^1), ..., (O^n, D^n)) = (d_x,d_y,d_z),
\end{equation}
where the network $F$ takes the predicted 2D orientation maps $\{{O}^i\}$, rendered depth maps $\{{D}^i\}$ and a 3D query point $X$ as input to predict its 3D hybrid vector $d$.

Following~\cite{zheng2023hairstep}, we stack $\{O^i\}$ and $\{D^i\}$ of input view $i$ channel-wisely and feed them to a backbone network $H$~\cite{newell2016stacked} to produce a deep feature map $f_i= H(O^i, D^i)$.
We then fetch pixel-aligned features $f_i(x)$ of 3D query point $X$ from each view. 
Then, we concatenate the features from different sources together with the positional encoding~\cite{mildenhall2021nerf} of $X$, and feed them into the decoder to predict the hybrid field value at position $X$.

The whole network is trained with the average $L_1$ loss of vector components on each axis. Specifically, we have:
\begin{equation}
L_{\text{ori}} = \frac{1}{N} \sum_{i}^{N} \frac{ \left\| d^* - d \right\|_1 }{3}.
\end{equation}
\vspace{-2em}





\paragraph{Strand growing}
Similar to the prior method~\cite{zheng2023hairstep}, directly growing hair strands $\{S^i\}$ from the scalp roots in our hybrid field is fast but not always complete.
When the predicted orientations are slightly inaccurate, the growth direction may deviate near the boundary, causing strands to stop prematurely and resulting in missing tails or unreachable regions.
In contrast, the segment-based strategy of~\cite{wu2024monohair} grows numerous short strand segments from random seeds within the occupancy field and then connects them using robust geometric rules, yielding highly accurate yet computationally expensive results for full hairstyles.

We observe that scalp-rooted growth already covers the majority of visible regions, leaving only small gaps.
Therefore, we adopt a hybrid growing strategy~\cite{chai2013dynamic}: we first grow the main strands from the scalp, and then generate supplementary short segments only in the missing areas to fill incomplete regions.
We attach each supplementary segment to the nearest scalp strand by appending all the preceding points of the closest strand and smoothing the connection for continuity.
This lightweight correction adds negligible computational cost while producing more complete and visually coherent hair reconstructions.

Some pre-sampled roots on the canonical head produce redundant or collapsed short strands.
To preserve valid buzz-cut hairs while keeping the overall hair clean, we keep only short strands projected inside the hair masks of at least two views.
As the reconstructed hybrid fields may be inaccurate in the buzz-cut region due to its extremely thin geometric structure, the generated buzz-cut hair strands tend to be sparse, and some strands collapse near their roots. 
To address this issue, we further recover the collapsed roots by reprojecting them into visible views, estimating the mean 3D orientation tangent to the mesh surface across multiple views, and regrowing the strands along the recovered direction.




\section{Experiment}
\label{sec:exp}

\subsection{Datasets}
To train the LoRA module, we construct a high-quality synthetic multi-view hair dataset.
Specifically, representative 3D hair strand models are carefully selected from USC-HairSalon~\cite{hu2015single} and Difflocks~\cite{rosu2025difflocks}.
Each model is registered to a template body mesh and rendered from 97 viewpoints along an orbital path under a static studio lighting setup to form 24 fps videos, with color augmentations for diversity.
An additional 100 3D models are randomly selected and rendered for evaluation.

For training the 3D hybrid field, we use 543 hair models in total, including 200 short and curly models from Difflocks and 343 models from USC-HairSalon, together with their mirrored versions with respect to the Y–Z plane.
We reserve 5\% of the data for evaluation.

To train the orientation extractor, we use 1,250 annotated frontal-view images and apply geometric transformations (translation, scaling, flipping, and rotation) to augment the data following~\cite{zheng2023hairstep}.
Additionally, we manually annotate 395 frames from real-world multi-view hair capture videos~\cite{sklyarova2023neural,wu2024monohair}, covering a wide range of hairstyles, to evaluate full-view performance.

\subsection{Evaluation}
\label{subsec: evaluation}





\paragraph{Multi-view generation}
We evaluate the effectiveness of our multi-view generation pipeline on a selected synthetic dataset, comprising 50 samples from USC-HairSalon\cite{hu2015single} and 50 from Difflocks\cite{rosu2025difflocks}.
We first measure the difference between the synthesized novel-view hair image and the rendering of the ground-truth strands using four metrics: $L_1$ error, PSNR, LPIPS\cite{zhang2018unreasonable}. We also compute the CLIP similarity\cite{liu2023one,tang2023dreamgaussian} between the generated views and the input view. The quantitative results are \textbf{0.049} for $L_1$, \textbf{20.897} for PSNR, and \textbf{0.162} for LPIPS. These values demonstrate that our generative module synthesizes perceptually reasonable novel-view hair. Furthermore, a CLIP similarity score of \textbf{0.834} indicates high perceptual consistency between the synthesized view and the input single view.
Please check visual results in the supplementary material.

\paragraph{3D hybrid field reconstruction}
Although we only predict the hybrid field, the occupancy can be easily inferred by checking whether the magnitude of the orientation vector at a query point is sufficiently small (we set a threshold of 0.1 to determine the outside region), since the orientations inside the volume are learned with unit length.
To evaluate the predicted fields, we follow the metrics used in~\cite{zheng2023hairstep}.
For orientation accuracy, we compute the mean squared error between the predicted and ground-truth orientations over 10k sampled points, achieving a value of \textbf{0.018}.
For occupancy accuracy, we calculate the IoU and precision, obtaining \textbf{0.986} and \textbf{0.997}, respectively.

\begin{table}[t]
\centering
\footnotesize
\caption{Quantitative comparisons of full-view orientation extraction (mean angle error, lower is better). \#Views means the number of annotated views.}
\label{tab:compare_orien_transposed}
\begin{tabular}{l| r| r| r| r| r| r| r| r}
\toprule

& nastya 
& ksyusha 
& jenya 
& shortCurly 
& whiteCurly 
& midWavy 
& midCurly 
& Overall \\
\midrule
\#Views 
& 62 & 66 & 70 & 58 & 42 & 33 & 64 & \textbf{395} \\

Gabor 
& 11.88 & 16.56 & 18.88 & 18.87 & 24.56 & 11.31 & 10.79 & \textbf{16.05} \\

HairStep 
& 6.28 & 10.98 & 13.16 & 13.68 & 21.57 & 6.32 & 5.54 & \textbf{10.88} \\

Ours 
& \textbf{1.79} 
& \textbf{4.97} 
& \textbf{6.09} 
& \textbf{6.72} 
& \textbf{13.50} 
& \textbf{2.15} 
& \textbf{2.12} 
& \textbf{5.13} \\
\bottomrule
\end{tabular}
\end{table}
\begin{figure}
	\centering
	\includegraphics[width=1.0\linewidth]{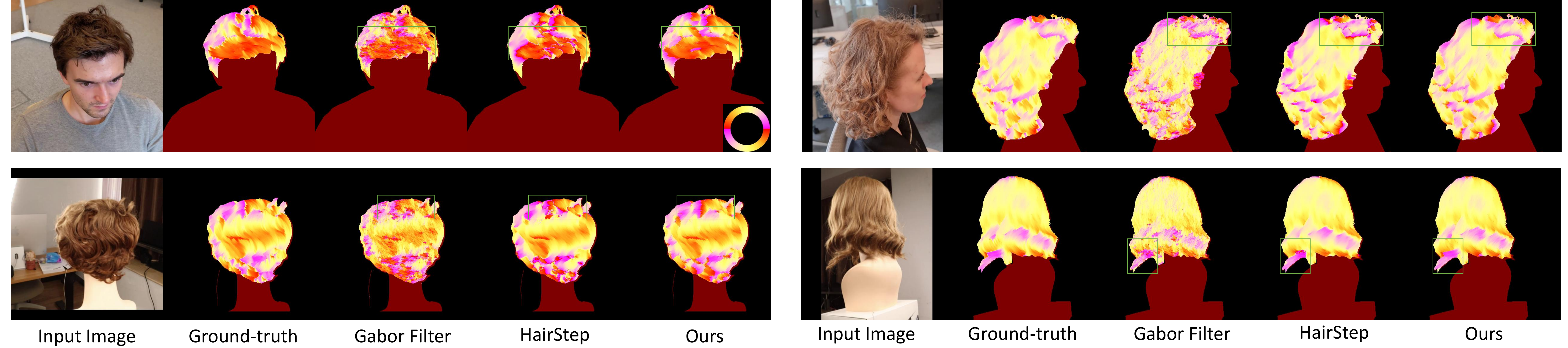}
	\caption{Qualitative comparisons of full-view orientation extraction. Results of HairStep have been converted to orientation maps.}
    \label{fig:comp_orien}
    \vspace{-1em}
\end{figure}

\paragraph{Full-view orientation extraction}
Furthermore, we evaluate the performance of our orientation extraction module on the annotated full-view dataset containing 395 images, and compare it against the Gabor filter~\cite{zhou2018hairnet} and the strand map from HairStep~\cite{zheng2023hairstep} (converted into an orientation map for fair comparison).
We compute the mean angular error between the predicted and ground-truth orientation maps over all pixels within the hair mask region.
As shown in Tab.~\ref{tab:compare_orien_transposed}, our extracted orientations significantly outperform both the Gabor filter and the orientation derived from HairStep.
A visual comparison is provided in Fig.~\ref{fig:comp_orien}.
While we employ the same network architecture as~\cite{zheng2023hairstep}, our key difference lies in redefining the learning target from a directional strand map to an undirectional orientation map. This reformulation enables more accurate and generalizable full-view orientation estimation, overcoming the frontal-view limitation of HairStep~\cite{zheng2023hairstep}. It opens the possibility of removing the reliance of hair modeling on the noisy outputs generated by Gabor filters.

\subsection{Comparison}

\begin{figure}
	\centering
	\includegraphics[width=\linewidth]{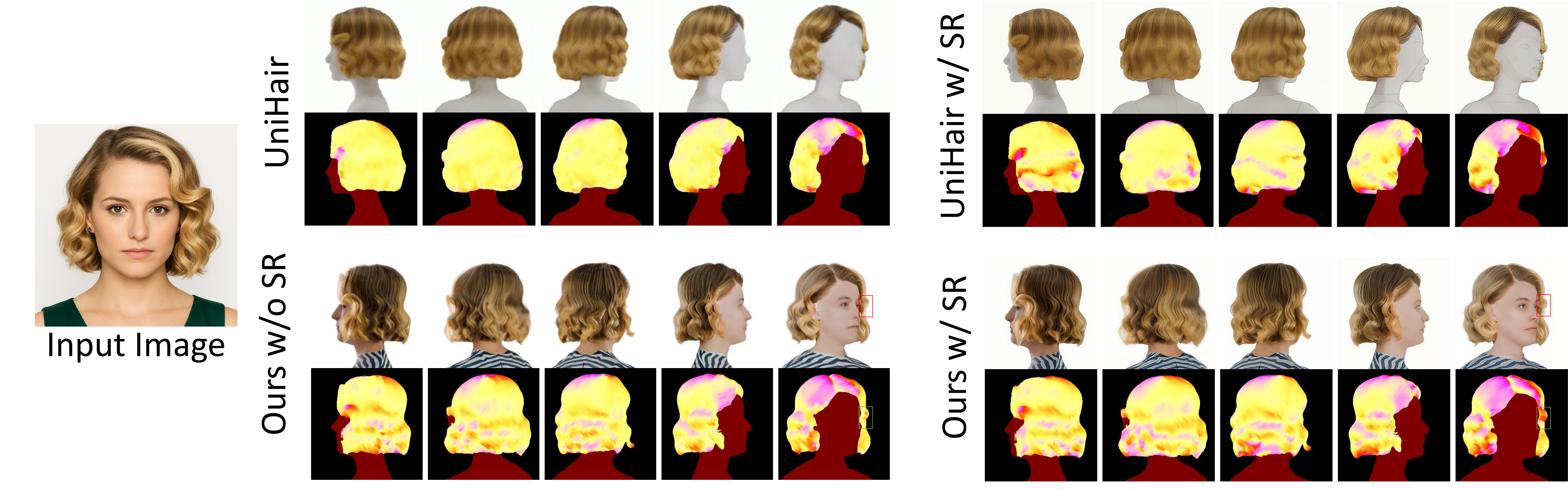}
	\caption{Comparisons on multi-view generation.}
    \label{fig:comp_unihair}
    \vspace{-1em}
\end{figure}

\begin{figure*}
	\centering
	\includegraphics[width=0.9\linewidth]{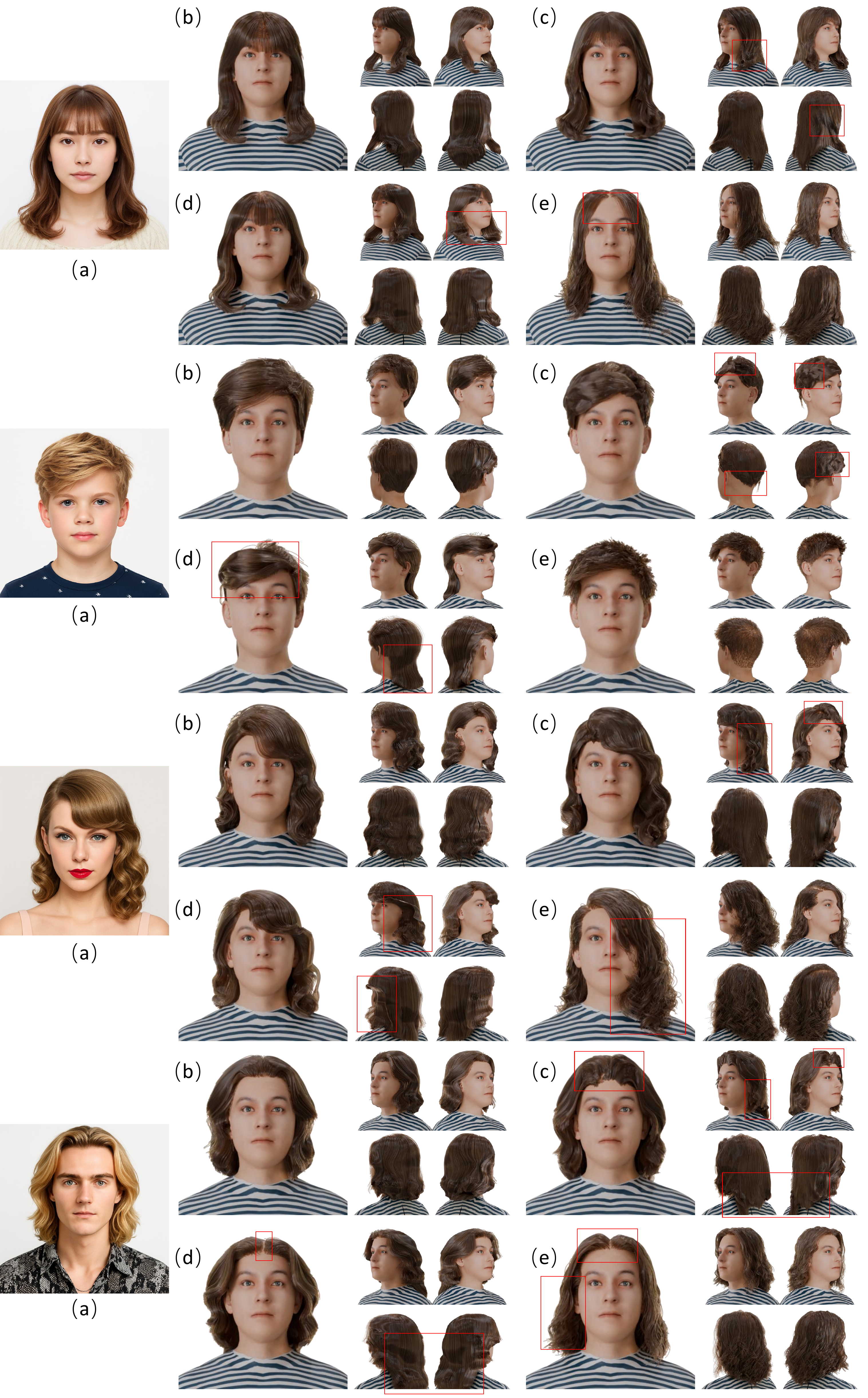}
	\caption{Qualitative comparison on single-view 3D strands reconstruction. For every example, we show (a) the input image and the reconstructed 3D hair strands rendered in multiple views of (b) Ours, (c) Im2Haircut, (d) HairStep and (e) Difflocks. }
    \label{fig:comp_svr}
    \vspace{-1em}
\end{figure*}


To comprehensively assess the performance of \methodName, we conduct comparisons on multi-view generation and single-view 3D strands reconstruction, respectively.

\paragraph{Comparison on multi-view generation}
A recent work, UniHair~\cite{zheng2024towards}, also attempts to generate multi-view hair images using a diffusion prior fine-tuned on synthetic data with both view-level and pixel-level Gaussian refinements.
We compare our multi-view generation module against UniHair and its enhanced version, UniHair-SR, which incorporates a post-processing step using the same super-resolution we adopt. 
As illustrated in~\cref{fig:comp_unihair}, all methods are capable of generating perceptually reasonable surrounding views from a single input hair image.
However, our results exhibit richer real-world hair texture details and are more consistent in style with the input image.
In contrast, the results of UniHair appear more synthetic.
Furthermore, the hair images generated by UniHair suffer from severe blurring, resulting in inaccurate orientation prediction.
Even after super-resolution enhancement, UniHair-SR produces unnatural strand details, which are detrimental to subsequent orientation extraction.
In comparison, our multi-view generation module produces high-quality surrounding-view images, significantly improving the performance of orientation extraction.
The extracted orientation maps not only capture global hairstyle characteristics, such as overall curl patterns, but also maintain smooth and continuous local details.
More results are provided in the supplementary.

\paragraph{Comparison on single-view 3D strands reconstruction}
We compare our method with recent works HairStep\cite{zheng2023hairstep}, Im2Haircut\cite{sklyarova2025im2haircut}, and Difflocks\cite{rosu2025difflocks}, the state-of-the-art methods on single-view 3D hair strands reconstruction. 
To compare quantitatively, we follow~\cite{zheng2023hairstep} to adopt HairSale, HairRida and IoU as the evaluation metrics to assess the single-view alignment by projecting the reconstructed 3D strands onto the image plane to calculate the angle error, accuracy of relative depth and IoU against manually annotated 196 hairstyles. 
From the quantitative results shown in~\cref{tab:compare_svr}, we can see our method outperforms on both metrics significantly.

Visual comparisons are also provided in~\cref{fig:comp_svr}. 
Im2Haircut and HairStep demonstrate good overall alignment with the input image, benefiting from a cleaner strand map and differentiable rendering–based post-optimization, respectively.
However, both methods fail to reconstruct the invisible regions, such as the side and back views, where the length and curliness of the hair are inconsistent with those in the input front view.
Moreover, the relative depth and the results obtained from DepthPro~\cite{bochkovskii2024depth}, which they rely on, are not sufficiently accurate.
Consequently, their reconstructions exhibit messy and unsmooth strand geometries, particularly around the frontal–side transition region.
Difflocks generates 3D hair strands that appear to match the hairstyle in the input image at first glance.
However, it fails to capture the actual hairstyle characteristics presented in the input view, such as the frontal bangs in the first example and the distinct curls and bends in the third and fourth examples.
A potential reason is that Difflocks directly generates a scalp texture latent map from the DINOv2~\cite{oquab2023dinov2} features of the input image through a diffusion process, without establishing sufficient spatial correspondence with the original image.
In contrast, our method generates the most accurate and multi-view plausible 3D hair strands compared with previous SOTA methods.


\begin{table}
\centering
\footnotesize
\caption{Quantitative comparisons on single-view 3D strands reconstruction with HairSale and IoU.}
\label{tab:compare_svr}
\begin{tabular}{lcccc}
\toprule
\textbf{Metric} & \textbf{HairStep} & \textbf{DiffLocks} & \textbf{Im2Haircut} & \textbf{Ours} \\
\midrule
HairSale ($^\circ$) $\downarrow$ & 17.38 & 26.50 & 16.24 & \textbf{12.83} \\
HairRida ($\%$) $\uparrow$ & 77.01 & 69.67 & 79.38 & \textbf{80.52} \\
IoU $\uparrow$ & 0.639 & 0.593 & 0.757 & \textbf{0.847} \\

\bottomrule
\end{tabular}
\vspace{-1em}
\end{table}

\subsection{Ablation} 




To comprehensively investigate the effectiveness of each design within our pipeline, we conduct an ablation study with the following four settings:

\begin{itemize}
\item $\bm{C_0}$: Replace our full-view orientation extractor with the traditional Gabor filter.
\item $\bm{C_1}$: Modify the strand growing method to purely grow strands from the scalp.
\item $\bm{C_2}$: Reduce the conditional views used for 3D field inference to a single input view only (i.e., disables our multi-view bridging strategy).
\item \textbf{\textit{Full}}: Complete \methodName pipeline.
\end{itemize}

\vspace{-2em}

\begin{figure*}
	\centering
	\includegraphics[width=0.88\linewidth]{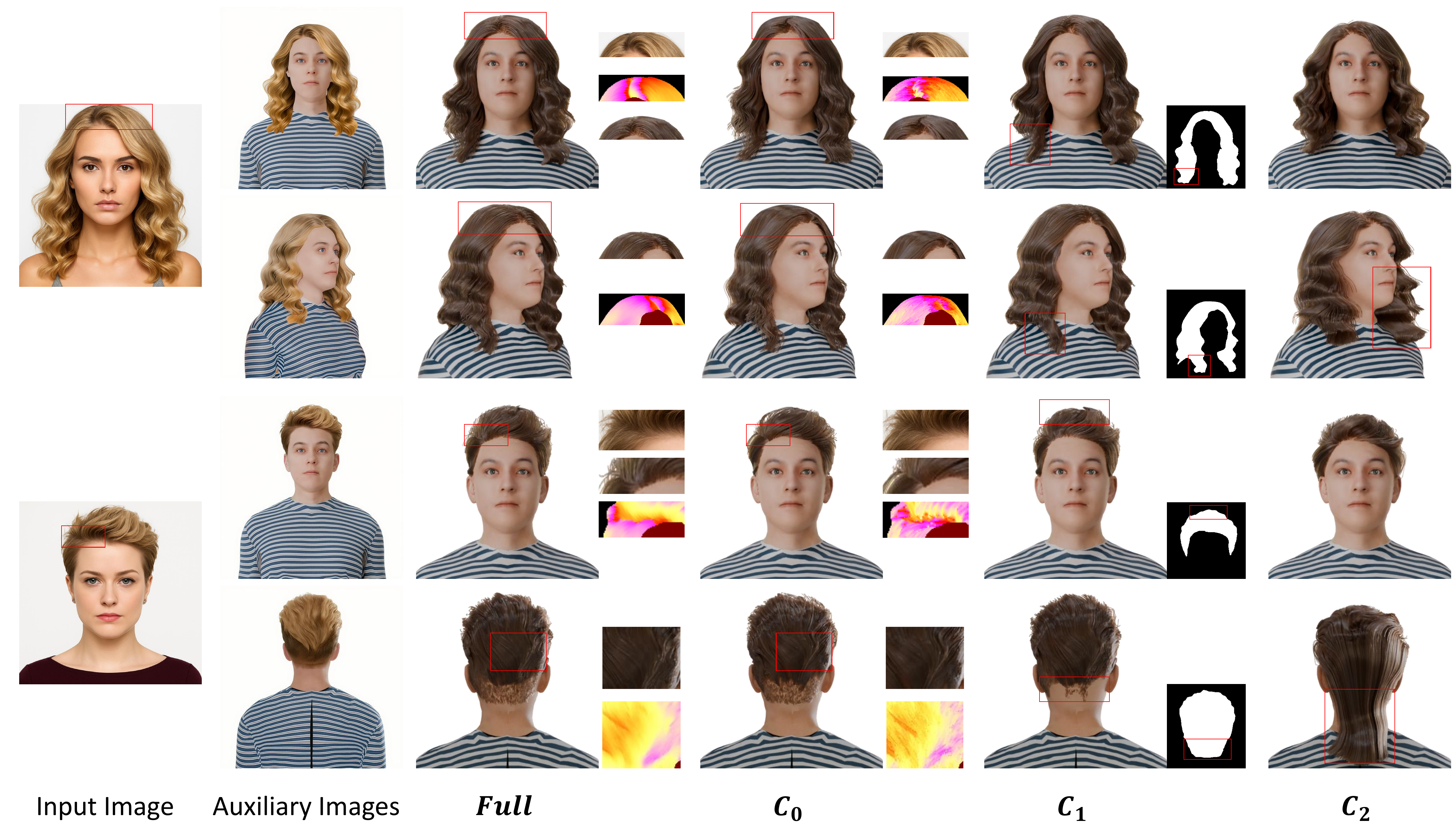}
	\caption{Qualitative comparisons of ablation study.}
    \label{fig:abla}
    \vspace{-1em}
\end{figure*}
\begin{table}
\centering
\footnotesize
\caption{Quantitative comparisons of ablation study. \textcolor{blue}{Blue} values denote the performance degradation (\%) compared to  \textbf{\textit{Full}}.}
\label{tab:ablation}
\begin{tabular}{lcc|cc}
\toprule
\multirow{2}{*}{\textbf{Method}} & \multicolumn{2}{c|}{\textbf{@Input View}} & \multicolumn{2}{c}{\textbf{@Synthesized 8 Views}} \\
\cmidrule(lr){2-3} \cmidrule(lr){4-5}
 & HairSale ($^\circ$) ↓ & IoU ↑ & HairSale ($^\circ$) ↓ & IoU ↑ \\
\midrule

$\bm{C_0}$      & 16.42 \textcolor{blue}{(-27.98\%)} & 0.794 & 13.45 \textcolor{blue}{(-34.90\%)} & 0.878 \\
$\bm{C_1}$   & 13.70 \textcolor{blue}{(-6.78\%)} & 0.828 & 10.37 \textcolor{blue}{(-4.01\%)} & 0.887 \\
$\bm{C_2}$   & 19.10 \textcolor{blue}{(-48.87\%)} & 0.680 & 24.14 \textcolor{blue}{(-142.1\%)} & 0.658 \\
\textbf{\textit{Full}} & \textbf{12.83} & \textbf{0.847} & \textbf{9.97} & \textbf{0.900} \\
\bottomrule
\end{tabular}
\vspace{-1em}
\end{table}

For each configuration, we report the quantitative results of HairSale and IoU for both the input view and the average across 8 synthetic novel views in~\cref{tab:ablation}.
We also provide visual illustration in~\cref{fig:abla}.

\paragraph{Ablation on full-view orientation extractor}
Compared with $\bm{C_0}$, our proposed orientation extractor significantly boosts the reconstruction accuracy on the input view and on the synthesized multiple views (\cref{tab:ablation}). 
This substantial improvement indicates that our orientation extractor infers more accurate and view-consistent hair orientations than the traditional Gabor filter. 
The visual comparisons between \textbf{\textit{Full}} and $\bm{C_0}$ shown in~\cref{fig:abla} further validate these observations.
The orientation maps derived from image filters often contain severe noise, which particularly affects key features such as hair partitioning during reconstruction. 

\paragraph{Ablation on the strand growing algorithm}
Compared with $\bm{C_1}$, our strand growing strategy achieves a better performance (\cref{tab:ablation}). 
The modest improvement indicates that scalp-rooted growth already covers the majority of the hair volume.
The visual results in~\cref{fig:abla} demonstrate that our method reconstructs more accurate strand geometry, particularly yielding more complete structures in the hair boundary and buzz-cut regions.

\paragraph{Ablation of multi-view bridging strategy}
Compared with $\bm{C_2}$, the incorporation of our multi-view bridging strategy yields a significant improvement for both the input view and the synthesized multiple views (\cref{tab:ablation}). 
The visual comparisons between \textbf{\textit{Full}} and $\bm{C_2}$ in~\cref{fig:abla} clearly demonstrate the effectiveness of integrating multi-view constraints in the 3D field inference pipeline, enhancing both local detail and view consistency.
Without synthesizing multiple views, $\bm{C_2}$ cannot estimate an accurate depth map for the input view.
Therefore, we use relative depth following~\cite{zheng2023hairstep}.

\section{Conclusion}
\label{sec:conclusion}

In this work, we revisit the long-standing challenge of reconstructing strand-level 3D hair geometry from a single image and reformulate it as a calibrated multi-view reconstruction problem.
By leveraging the intrinsic cross-view priors encoded in video diffusion models, our approach synthesizes geometrically consistent and texture-rich multi-view observations from a single portrait.
We further propose a full-view neural orientation extractor, a hybrid implicit field representation, and an efficient strand-growing strategy that jointly enhance reconstruction fidelity, completeness, and efficiency.
Extensive experiments demonstrate the superiority of our method.
Limitations and future works are discussed in the supplementary.

%
%
\bibliographystyle{splncs04}
\bibliography{main}
\end{document}